\documentclass[letterpaper, 10 pt, conference]{ieeeconf}

\IEEEoverridecommandlockouts
\usepackage[utf8]{inputenc} 
\usepackage[T1]{fontenc}    
\usepackage{url}            
\usepackage{booktabs}       
\usepackage{amsfonts}       
\usepackage{nicefrac}       
\usepackage{microtype}      
\usepackage{xcolor}         

\usepackage{url}
\usepackage{graphicx}
\usepackage{caption} 
\usepackage{algorithm}
\usepackage[noend]{algpseudocode}
\usepackage{amsmath,amssymb}
\usepackage{multirow}
\usepackage{array,multirow,graphicx}
\usepackage{float}
\usepackage{wrapfig}
\usepackage{todonotes}
\usepackage{subcaption}
\usepackage{amsmath}
\def\etal{\emph{et al}.}

\newcommand{\bo}{\mathbf{o}}
\newcommand{\bs}{\mathbf{s}}
\newcommand{\ba}{\mathbf{a}}
\newcommand{\br}{\mathbf{r}}

\newcommand{\st}{\bs_t}
\newcommand{\at}{\ba_t}

\newcommand{\stp}{\bs_{t+1}}

\newcommand{\states}{\mathcal{S}}
\newcommand{\actions}{\mathcal{A}}

\newcommand{\qattn}{Q}

\newcommand{\qattnp}{\theta}
\newcommand{\qrotgripp}{\phi}

\newcommand{\obs}{\mathbf{o}}

\newcommand{\rgb}{\mathbf{b}}

\newcommand{\pcd}{\mathbf{p}}
\newcommand{\proprio}{\mathbf{z}}

\newcommand{\replay}{\mathcal{D}}

\DeclareMathOperator{\E}{\mathbb{E}}

\DeclareMathOperator*{\argmax}{arg\,max}

\DeclareMathOperator*{\argmaxthreed}{\argmax3D}

\newcommand{\qdepthN}{N}
\newcommand{\qdepth}{n}
\newcommand{\vox}{\mathbf{v}}
\newcommand{\voxf}{\mathbf{V}}
\newcommand{\centre}{\mathbf{c}}
\newcommand{\res}{\mathbf{e}}

\newcommand{\topklist}{\eta}

\newcommand{\methAcro}{QTE}

\definecolor{newCodeColor}{RGB}{194,218,194}

\graphicspath{{images/}}

\title{\LARGE \bf
Coarse-to-fine Q-attention with Tree Expansion
}

\author{Stephen James and Pieter Abbeel\\
UC Berkeley\\
\{stepjam, pabbeel\}@berkeley.edu
}

\begin{document}
\bstctlcite{IEEEexample:BSTcontrol}

\maketitle
\thispagestyle{plain}
\pagestyle{plain}

\begin{abstract}
Coarse-to-fine Q-attention enables sample-efficient robot manipulation by discretizing the translation space in a coarse-to-fine manner, where the resolution gradually increases at each layer in the hierarchy. Although effective, Q-attention suffers from ``coarse ambiguity'' --- when voxelization is significantly coarse, it is not feasible to distinguish similar-looking objects without first inspecting at a finer resolution. To combat this, we propose to envision Q-attention as a tree that can be expanded and used to accumulate value estimates across the top-k voxels at each Q-attention depth. When our extension, Q-attention with Tree Expansion (QTE), replaces standard Q-attention in the Attention-driven Robot Manipulation (ARM) system, we are able to accomplish a larger set of tasks; especially on those that suffer from ``coarse ambiguity''. In addition to evaluating our approach across 12 RLBench tasks, we also show that the improved performance is visible in a real-world task involving small objects. Videos and code found at: \url{https://sites.google.com/view/q-attention-qte}.
\end{abstract}

\section{Introduction}

Coarse-to-fine Q-attention~\cite{james2021coarse}, and its role within the Coarse-to-fine Attention-driven Robotic Manipulation (C2F-ARM) system, has increased the feasibility of training vision-based, sparse-rewarded reinforcement learning agents in the real world. Given a small number of initial demonstrations and quick exploration phase, C2F-ARM learns to output a series of next-best poses which when given to a motion planner, leads to successful completion of manipulation tasks. The algorithm gains its sample efficiency, in part, due to the discretization of the translation space (voxelization) via the coarse-to-fine Q-attention. To overcome the need to voxelize a large workspace --- which would be memory intensive and infeasible for a large number of voxels --- coarse-to-fine Q-attention gradually increases the resolution by extracting the highest valued voxel at each level in the hierarchy, and using it as the center of the higher-resolution voxelization in the next level. Because this coarse-to-fine prediction is applied over several (increasingly finer) levels, it gives a near-lossless prediction of the translation.

\begin{figure}
\centering
\includegraphics[width=1.0\linewidth]{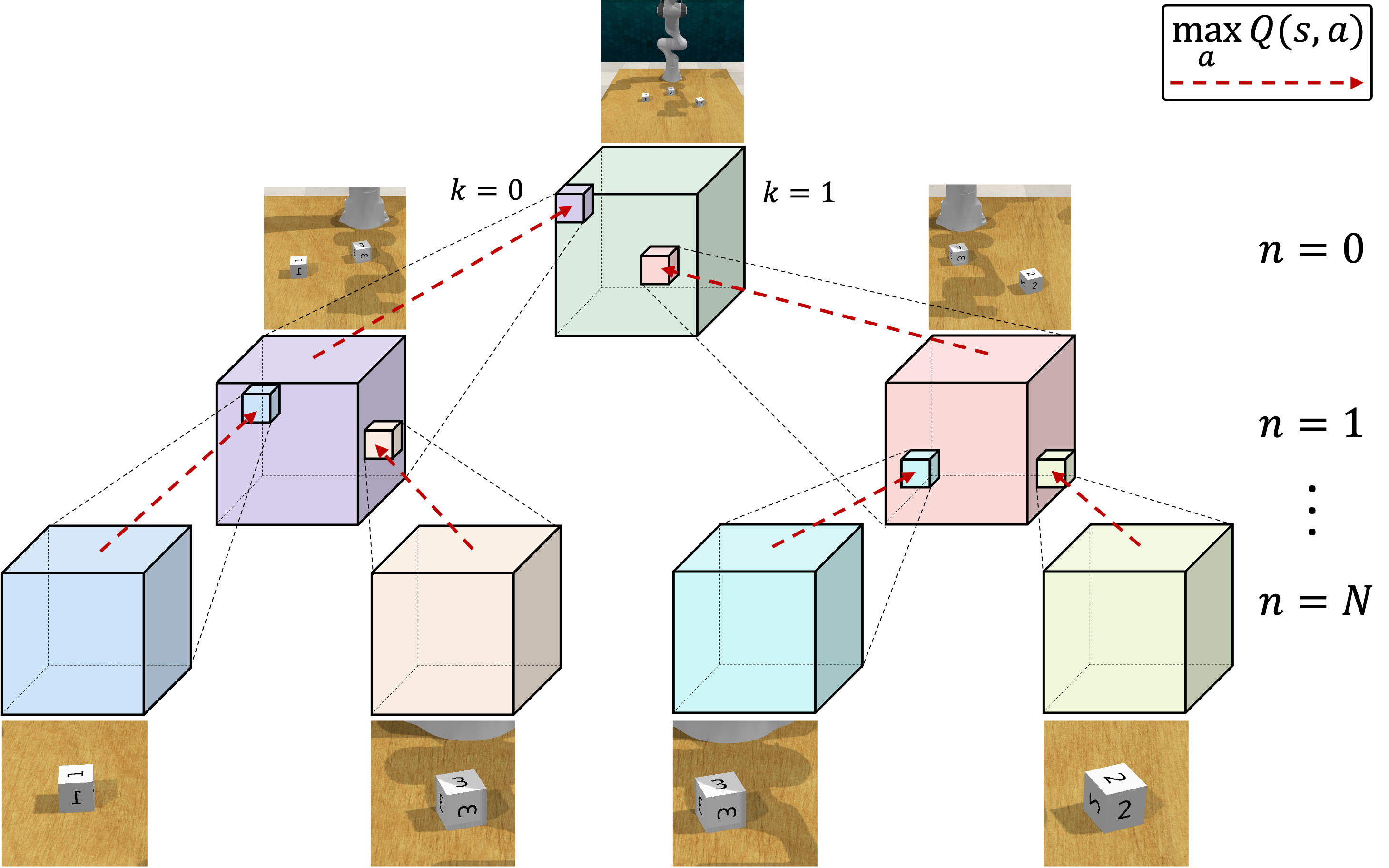}
\caption{At the top-level of the coarse-to-fine Q-attention, it is not feasible to distinguish similar-looking objects without first inspecting at a finer resolution. For example, at the coarsest voxelization, the numbers on the blocks above will not be legible. \textbf{Q-attention with Tree Expansion (QTE)} expands (dotted black lines) and accumulates value estimates (dashed red lines) across the top-k voxels at each Q-attention depth.}
\label{fig:method_summary}
\end{figure}

\begin{figure}
\centering
\includegraphics[width=1.0\linewidth]{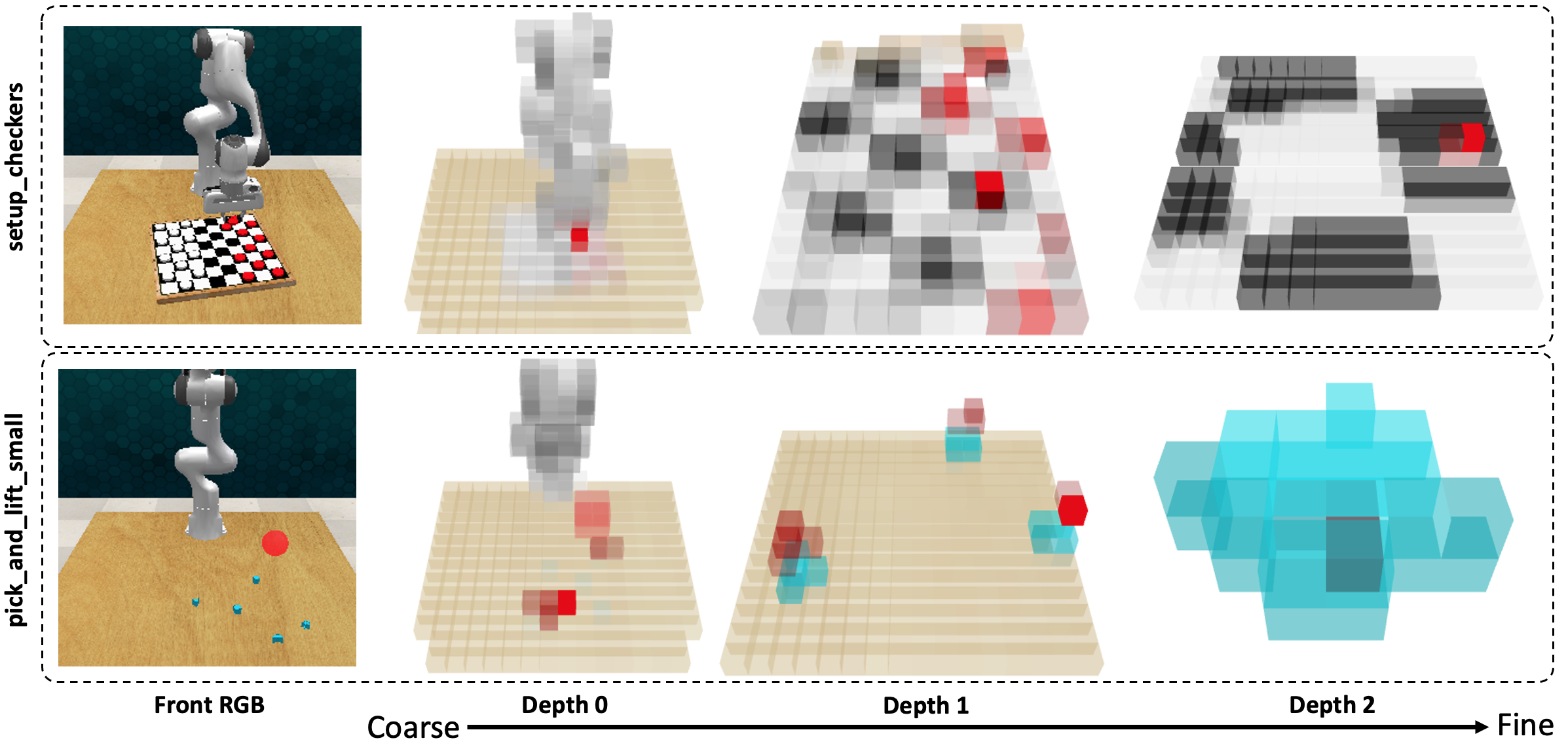}
\caption{Example of ``coarse ambiguity''. At the coarsest levels of the Q-attention, it is difficult to distinguish between objects. Highlighted red voxels show the attention.}
\label{fig:c2f_vis}
\end{figure}

Although effective in a number of simulation and real-world tasks, Q-attention suffers from what we call ``coarse ambiguity'' --- when the voxelization is significantly coarse, it is not feasible to distinguish between similar-looking objects without first inspecting them at a finer resolution. Take Figure \ref{fig:method_summary} as a motivating example, where the goal is to grasp the cube with the number `1'; at the top level voxelization, the numbers on the blocks will not be legible, thus limiting the root Q-attention's ability to make an informed decisions about which block to attend to. This ``coarse ambiguity'' is further shown in Figure \ref{fig:c2f_vis}.
Unfortunately, simply increasing the voxel grid size at the root of the Q-attention (i.e., directly operating at a finer resolution) would require significantly memory, and would not scale to larger workspaces; therefore an alternative solution is required.

In this work, we look to improve the performance of coarse-to-fine Q-attention by enriching the information available to the coarsest Q-attention layers. We do this by taking inspiration from Monte Carlo Tree Search (MCTS) --- where Monte Carlo simulation is used to accumulate value estimates to guide towards highly rewarding trajectories in a search tree. However, rather than searching across \textit{time}  (i.e., simulating a `game'), we instead search and accumulate value estimates across \textit{space}, thus allowing values from the fine-layers to propagate to the coarse layers and overcoming the ``coarse ambiguity''.

\begin{figure}
\centering
\includegraphics[width=1.0\linewidth]{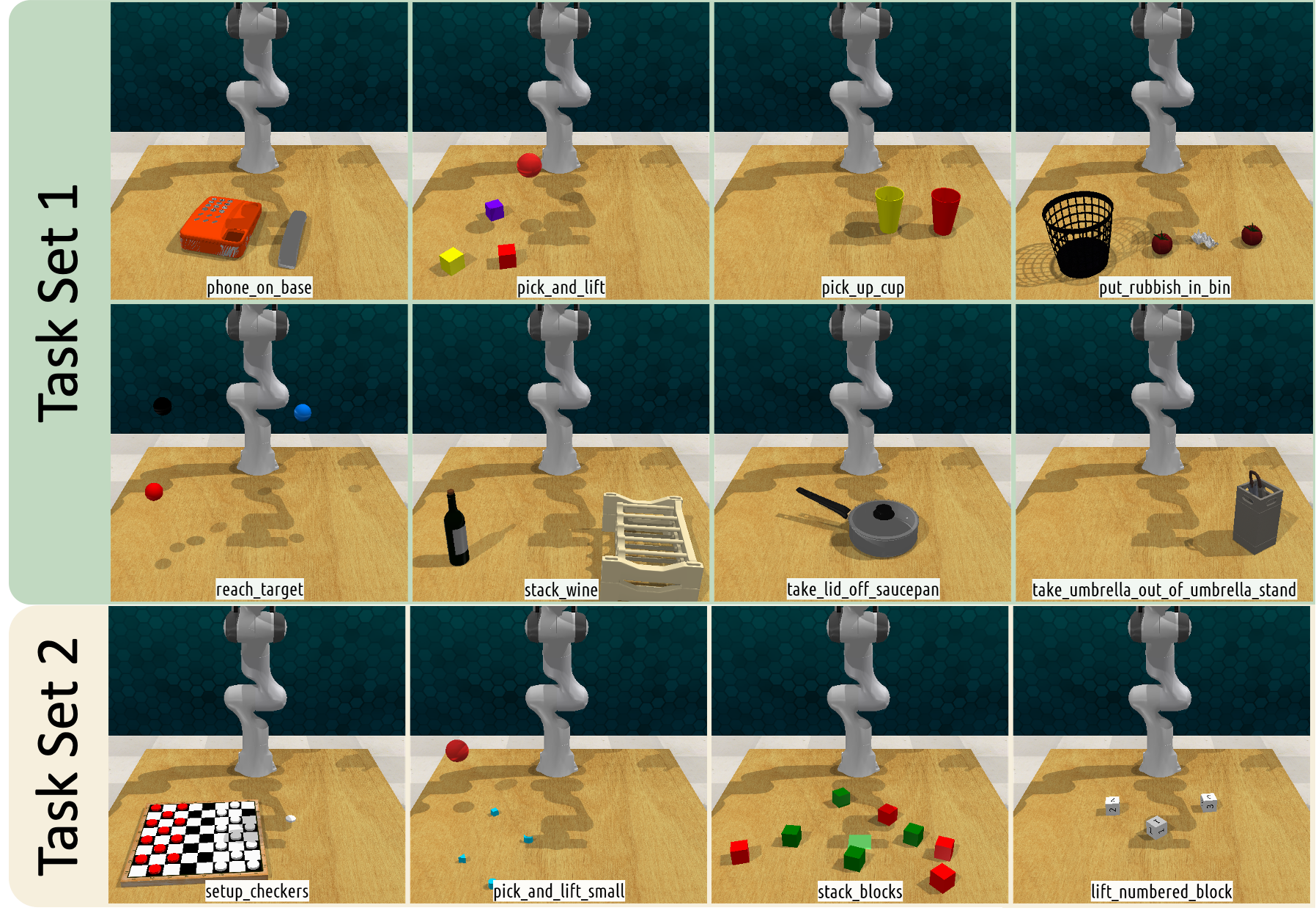}
\caption{In simulation, our method is evaluated on a total of 12 RLBench~\cite{james2019rlbench} tasks. \textit{Task Set 1} are tasks that have previously been shown to perform well with C2F-ARM. \textit{Task Set 2} are tasks that were chosen due to their ambiguity at a coarse scale. Note that the position and orientation of objects are placed randomly at the beginning of each episode.}
\label{fig:sim_taskset}
\end{figure}

Our method, Q-attention with Tree Expansion (QTE), is built into the Attention-driven Robot Manipulation (ARM) system to give our new C2F-ARM+\methAcro\ system. This system is evaluated on 12 simulated RLBench~\cite{james2019rlbench} tasks (Figure \ref{fig:sim_taskset}) and compared against vanilla C2F-ARM. We show that C2F-ARM+\methAcro\ can accomplish a wider range of tasks, and much like C2F-ARM, is capable of learning sparsely-rewarded real-world tasks.

To summarize, we propose our main contribution, Q-attention with Tree Expansion (QTE): an extension to coarse-to-fine Q-attention that allows top-level coarse layers to inspect the lower-level fine layers before action selection. Along with this contribution, we also propose a new system, C2F-ARM+\methAcro, which improves the performance over C2F-ARM on a set of RLBench tasks.

\section{Related Work}
\label{sec:related_work}

\textbf{Robot Manipulation} through the use of deep reinforcement learning (RL) has been used to solve a variety of tasks, including object pushing~\cite{pinto2017asymmetric}, block lifting~\cite{james20163d}, cloth manipulation~\cite{matas2018sim}, lego stacking~\cite{haarnoja2018composable}, and grasping~\cite{kalashnikov2018qt, james2019sim}, to name but a few. Rather than using continuous control actor-critic reinforcement learning algorithms, a recent trend instead discretizes part, or all, of the action space~\cite{morrison2018closing, zeng2018learning, zeng2020transporter, james2022qattention, james2021coarse}. In particular, the recently proposed C2F-ARM~\cite{james2021coarse} has shown the ability to rapidly learn sparse-reward, and image-based manipulation tasks by discretizes the action space in a coarse-to-fine manner via a Q-attention network. The voxel representation used in C2F-ARM~\cite{james2021coarse} has sporadically been used for manipulation. Notable manipulation works include MoreFusion~\cite{wada2020morefusion}, which used voxels to perform multi-object reasoning to improve 6D pose estimation for precise pick-and-place in clutter; it has also been used for safe object extraction~\cite{wada2022safepicking}, and object reorientation~\cite{wada2022reorientbot}. Song \etal~\cite{song2020grasping} and Breyer \etal~\cite{breyer2021volumetric} fed a voxel grid representation to a neural network to generate 6 DoF grasp actions. 

In this work, we aim to improve and extend the coarse-to-fine Q-attention component of ARM. However, this is not the first work to propose extensions to ARM. ARM~\cite{james2022qattention}, has recently been extended to a Bingham policy parameterization~\cite{james2022bingham} to improve training stability. Another extension has sought to improve the control agent within coarse-to-fine ARM to use learned path ranking~\cite{james2022lpr} to overcome the weaknesses of traditional path planning. We aim to further improve coarse-to-fine Q-attention and tackle the ``coarse ambiguity'' issue that Q-attention suffers from. Our solution shares motivational similarities with Monte Carlo Tree Search, which we cover below.

\textbf{Monte Carlo Tree Search} (MCTS) has predominantly been used for decision making within combinatorial games, including Go\cite{kocsis2006bandit}, Poker\cite{broeck2009monte}, and Chess\cite{silver2018general}, to name but a few. Driven by successes in games, MCTS has increasingly been applied outside of this area; one such case is within the realm of robot manipulation. 
Rearrangement planning --- where the goal is to find a sequence of transit and transfer motions to move a set of objects to a target arrangement --- has recently seen success from MCTS. \textit{Labbe et al.}~\cite{labbe2020monte} combined Monte Carlo Tree Search with a standard robot motion planning algorithm (e.g., RRT). \textit{Eljuri et al.}~\cite{eljuri2021combining} combined MCTS and a Motion Feasibility Checker (MFC), which selects a set of feasible poses for the robot from a feasibility database. Recently, MCTS has been combined with `Task and Motion Planning' (TAMP), by constructing an extended decision tree for symbolic task planning and high-dimension motion variable binding~\cite{ren2021extended}. 
Outside of symbolic planning, \textit{Huang et al.}~\cite{huang2021visual} learned to retrieving a target object from clutter performing a Monte Carlo search over high-level non-prehensile actions. Its search time was subsequently improved by learning to predict the discounted reward of each branch in MCTS without the need to roll out~\cite{huang2022self}. \textit{Bai et al.}~\cite{bai2021hierarchical} also tackle non-prehensile manipulation by proposing an MCTS algorithm guided by a policy network which is trained by imitation and reinforcement. 

Another related area that we briefly want to mention is \textbf{beam search} --- a search algorithm that expands the most promising nodes of a graph in a limited set. Beam search has almost exclusively been used in natural language processing, and to the best our knowledge, has not been explored for action selection in robotics.

\section{Background}
\label{sec:background}

\subsection{Reinforcement Learning}

The reinforcement learning problem consists of and agent and environment. The agents takes actions $\ba \in \actions$, and transitions to new states $\bs \in \states$, where $\st$ and $\at$ are the state and action at time step $t$. Upon transitioning to a new state, the agent receives a reward $r$ from a reward function $R(\st,\at)$. The goal of reinforcement learning is for the agent to learn a policy $\pi$ that aims to maximize the expected future (discounted) reward: $\E_\pi [\sum_t \gamma^t R(\st, \at)]$, where $\gamma \in [0, 1)$ is a discount factor. The expected return after taking action $\ba$ in state $\bs$ can be estimated via a value function $Q(s, a)$.

Deep Q-learning~\cite{mnih2015human} approximates the value function $Q_\psi$ via a deep convolutions network with parameters $\psi$. Exploration data is stored into a replay buffer, and subsequently used for sampling mini-batches of transitions to minimize: $\E_{(\st, \at, \stp) \sim \replay} [ (\br + \gamma \max_{\ba'}Q_{\psi'}(\stp, \ba') - Q_{\psi}(\st, \at))^2]$, where $Q_{\psi'}$ is a target network --- a periodic copy of $Q_\psi$, which helps prevent Q-value overestimation. The Q-attention module, which we discuss in the next section, builds upon the work of Deep Q-learning. 

\subsection{Coarse-to-fine Q-attention}
\label{sec:back:arm}

\begin{algorithm}[tb]
\caption{C2F-ARM+\methAcro. Addition to C2F-ARM in green.}
\label{alg:arm}
\begin{algorithmic}[1]
    \State Initialize Coarse-to-fine Q-attention $\qattn$.
    \State Initialize buffer $\replay$ with demos; 
    \State apply \textbf{keyframe discovery} and \textbf{demo augmentation}.

    \For{each iteration}
	    \For{each environment step $t$}
	        \State $\obs_t \leftarrow (\rgb_t, \pcd_t, \proprio_t)$
	        
	        \State $\centre^0 \leftarrow$ Scene centroid
	        \State $\text{coords} \leftarrow [\text{ }]$

	        \For{each ($\qdepth$ of $\qdepthN$) Q-attention depths}
	        
	            \State $\vox^\qdepth \leftarrow \voxf(\obs_t, \res^\qdepth, \centre^\qdepth)$
	            
    	        \State {\hspace*{-\fboxsep}\colorbox{newCodeColor}{\parbox{0.7\linewidth}{%
    	        $q^\qdepth \leftarrow \text{QTE}(\qdepth, \obs_t, \res^{\qdepth}, \centre^{\qdepth})$
    	        }}}
    	        
    	        \State {\hspace*{-\fboxsep}\colorbox{newCodeColor}{\parbox{0.7\linewidth}{%
    	        $(x^\qdepth, y^\qdepth, z^\qdepth) \leftarrow$ retrieve indices of $q^\qdepth$ 
    	        }}}
    	        
    	        \State $\text{coords.append(} (x^\qdepth, y^\qdepth, z^\qdepth) \text{)}$
    	        
    	       \If{$\qdepth == N$}
    	            \State $\alpha, \beta, \gamma, \omega \leftarrow \argmax_{\ba^h} \qattn^{h}_{\qrotgripp}(\tilde{\vox^\qdepthN}, \ba^h)$ for $h \in \{0, 1, 2, 3\}$
    	       \EndIf
    	        
    	        \State $\centre^{\qdepth+1} \leftarrow (x^\qdepth, y^\qdepth, z^\qdepth)$
	        \EndFor
	        
	        \State $\at \leftarrow (\centre^{\qdepthN}, \alpha, \beta, \gamma, \omega)$
	        \State $\obs_{t+1}, \br \leftarrow env.step(\at)$
            \State $\mathcal{D} \leftarrow \replay \cup \left\{(\obs_t, \at, \br, \obs_{t+1}, \text{coords})\right\}$
    	\EndFor
    	\For{each gradient step}
    	    \State $\qattnp_\qdepth \leftarrow \qattnp_\qdepth - \lambda_{\qattn} \hat \nabla_{\qattnp_\qdepth} J_{\qattn}(\qattnp_\qdepth)$ for $\qdepth \in \{0, \dots, \qdepthN\}$
    	    \State $\qattnp'_\qdepth \leftarrow \tau \qattnp_\qdepth + (1-\tau) \qattnp'_\qdepth$ for $\qdepth \in\{0, \dots, \qdepthN\}$
	    \EndFor
    \EndFor
\end{algorithmic}
\end{algorithm}

Coarse-to-fine Q-attention~\cite{james2021coarse} is a reinforcement learning (robot manipulation) module that accepts a coarse 3D voxelization of the scene, and learns to attend to salient areas (or objects) in the scene that would lead to task success. Upon choosing a salient area, it `zooms' into that area and re-voxelizes the scene at a a higher spacial resolution. By applying this coarse-to-fine process iteratively, we get a near-lossless discretization of the translation space. This is also combined with a discretized rotation prediction to give a next-best pose. This module was built into the C2F-ARM~\cite{james2021coarse} system, and resulted in the ability to learn sparsely-rewarded, image-based manipulation tasks in very few environment steps. The C2F-ARM system takes the next-best pose output from Q-attention, and uses a motion planner to get the actions required to reach the goal pose. Coarse-to-fine Q-attention is an extension of the earlier 2D Q-attention~\cite{james2022qattention} which attended to pixels, rather that voxels, and did not recursively apply the attention in a coarse-to-fine manner. 

In order to overcome exploration in a sparse reward environment, ARM~\cite{james2022qattention} (the predecessor to C2F-ARM) used a handful of demonstrations along with two demonstration pre-processing steps: keyframe discovery and demo augmentation. The \textit{Keyframe discovery} method aims to assist the the Q-attention to quickly converge and suggest meaningful points of interest in the initial phase of training; while \textit{demo augmentation} is a augmentation strategy that slices the demonstration trajectories into multiple smaller trajectories, thereby increases the initial number of demo transitions in the replay buffer.

\section{Method}
\label{sec:method}

\begin{algorithm}[tb]
\caption{Q-attention Tree Expansion (QTE).}
\label{alg:cte}
\begin{algorithmic}[1]

    \Procedure{QTE}{$\qdepth, \obs_t, \res^{\qdepth}, \centre^{\qdepth}$}
    
        \State $\vox^{\qdepth} \leftarrow \voxf(\obs_t, \res^{\qdepth}, \centre^{\qdepth})$
    
        \If{$\qdepth == \qdepthN$}  \Comment{Base case}
            \State \Return {$\max_{\ba'} \qattn_{\qattnp_{\qdepth}}(\vox^\qdepth, \ba')$}
        \Else \Comment{Recursive case}
    
            \State $\topklist \leftarrow \text{TOPK}_{\ba'}(\qattn_{\qattnp_{\qdepth}}(\vox^\qdepth, \ba'))$
            
            \State $\text{values} \leftarrow [\text{ }]$
            
            \For{$k$ in $\{0, \dots, K\}$}
                \State $q^k \leftarrow \topklist[k]$
                \State $(x^k, y^k, z^k) \leftarrow$ retrieve indices of $q^k$ 
                \State $\centre^{\qdepth+1} \leftarrow (x^k, y^k, z^k)$
                \State $q' = \text{QTE}(\qdepth+1, \obs_t, \res^{\qdepth+1}, \centre^{\qdepth+1})$
                \State $\text{values.append}((q^k + q')/2)$
            \EndFor
            \State \Return {$\max\text{values}$}
        \EndIf
    \EndProcedure
\end{algorithmic}
\end{algorithm}

Our addition to Q-attention is simple: we frame Q-attention as a tree that can be expanded and used to accumulate value estimates across the top-k voxels at each Q-attention depth. This is somewhat akin to (though not the same as) Monte Carlo Tree Search (MCTS) --- where Monte Carlo simulation is used to accumulate value estimates to guide towards highly rewarding trajectories in a search tree. Although similar in spirit, our Q-attention with Tree Expansion (\methAcro) expands and accumulates value estimates across \textit{space}, rather than \textit{searching} across \textit{time} --- i.e., we do not simulate a rollout in the environment.
Note that an obvious alternative to our approach would be to simply increase the number of voxels at the root of the Q-attention, however this would require significantly more memory, and would not be a scalable solution for larger workspaces. 
Another obvious alternative to tackle ``coarse ambiguity'' would be to simply operate on raw RGB-D images (rather than voxelization), however these raw-observation policies do not perform as well as C2F-ARM~\cite{james2022qattention}.

\begin{figure*}
\centering
\includegraphics[width=1.0\linewidth]{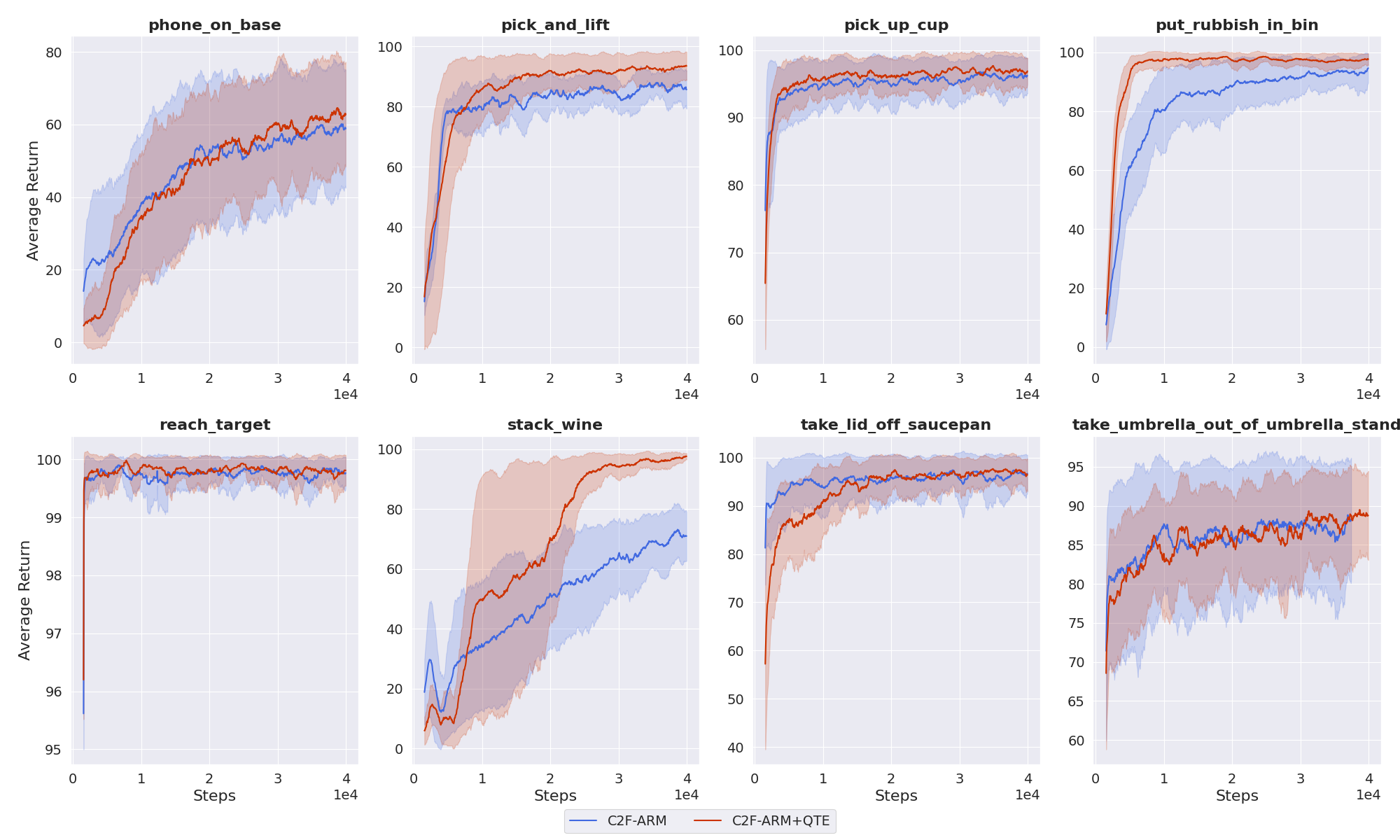}
\caption{Learning curves for 8 RLBench tasks. Both C2F-ARM and C2F-ARM+QTE only receive 10 demos, and are stored in the replay buffer prior to exploration. Note, these results intend to show that there is no loss in performance when QTE is not necessary, and we do not expect to outperform C2F-ARM. In Figure \ref{fig:results_new_tasks} however, the goal is to outperform C2F-ARM. Solid lines represent the average evaluation over 5 seeds, while the shaded regions represent the $std$.}
\label{fig:results_original_tasks}
\end{figure*}

\subsection{C2F-ARM+\methAcro}

Our system assumes we are operating in a partially observable Markov decision process (POMDP), where an observation $\obs$ consists of an RGB image, $\rgb$, an organized point cloud, $\pcd$, and proprioceptive data, $\proprio$. Actions consist of a 6D (next-best) pose and gripper action, and the reward function is sparse, giving $100$ on task completion, and $0$ for all other transitions.

We first formally define the coarse-to-fine Q-attention. We assume access to a voxelization function $\vox^\qdepth = \voxf(\bo, \res^\qdepth, \centre^\qdepth)$, which takes the observation $\obs$, a voxel resolution $\res^\qdepth$, and a voxel grid center $\centre^\qdepth$, and returns a voxel grid $\vox^{\qdepth} \in \mathbb{R}^{x×y×z×(3+M+1)}$ at depth $\qdepth$, where $\qdepth$ is the depth/iteration of the Q-attention, and where each voxel contains the 3D coordinates, $M$ features (e.g., RGB values, features, etc), and an occupancy flag.

Given our Q-attention function $\qattn_{\qattnp_{\qdepth}}$ at depth $\qdepth$, the original C2F-ARM system would extract the indices of the voxel with the highest value:
\begin{equation}
\label{eq:extractxyz}
\vox_{ijk}^{\qdepth} = \argmaxthreed_{\ba'} \qattn_{\qattnp_{\qdepth}}(\vox^\qdepth, \ba'),
\end{equation}
where $\vox_{ijk}$ is the extracted voxel index located at $(i, j, k)$, and $\argmaxthreed$ is an \textit{argmax} taken across three dimensions (depth, height, and width). However, we now extract the indices of the voxel with the highest value from the full tree expansion:
\begin{equation}
\label{eq:extractxyz_}
\vox_{ijk}^{\qdepth} = \text{extract\_index} ( \text{QTE}(\qdepth, \obs_t, \res^{\qdepth}, \centre^{\qdepth}) )
\end{equation}
where $\text{QTE}$ is the recursive tree expansion procedure discussed later in Section \ref{sec:tree_expansion}. Note that in practice, $\text{QTE}$ would also return the indices of the highest voxels, but these have been omitted for brevity. As laid out in \textit{James et al.}~\cite{james2022qattention}, we use a separate Q-network for each resolution. 

From this point, C2F-ARM proceeds as laid out in \textit{James et al.}~\cite{james2021coarse}. Using the extracted indices, $\vox_{ijk}^{\qdepth}$, we calculate the $(x^\qdepth, y^\qdepth, z^\qdepth)$ Cartesian coordinates of that voxel using known camera parameters. Given that the extracted coordinates represent the next-best coordinate to voxelize at a higher resolution, we can set these coordinates to be the voxel grid center $\centre$ for the next depth: $\centre^{\qdepth+1} = (x^\qdepth, y^\qdepth, z^\qdepth)$. However, if this is the last depth of the Q-attention, then $\centre^{\qdepthN} = \centre^{\qdepth+1}$ represents the continuous representation of the translation (i.e., the translation component of the next-best pose agent).

\begin{figure*}
\centering
\includegraphics[width=1.0\linewidth]{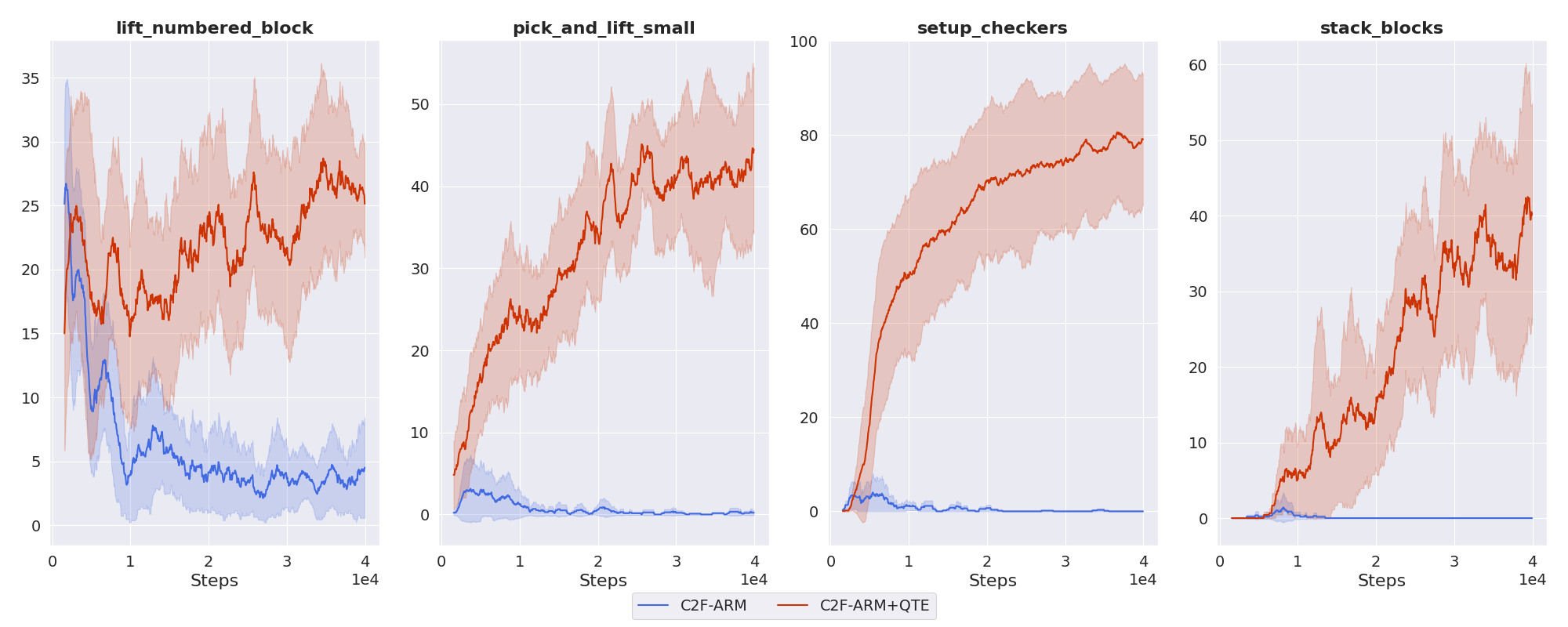}
\caption{Learning curves for an additional 4 RLBench tasks. Both methods receive 10 demos for \textit{lift\_numbered\_block} and \textit{pick\_and\_lift\_small}, 20 demos for \textit{setup\_checkers}, and 40 demos for \textit{stack\_blocks}. Unlike Figure \ref{fig:results_original_tasks}, we do expect C2F-ARM+QTE to outperform C2F-ARM, as these tasks cause ``coarse ambiguity''. Solid lines represent the average evaluation over 5 seeds, while the shaded regions represent the $std$.}
\label{fig:results_new_tasks}
\end{figure*}

As done in \textit{James et al.}~\cite{james2021coarse}, the rotation of each axis is discretized in increments of $5$ degrees, while the gripper is discretized to an open or close binary. These components are recovered from an MLP branch (with parameters $\qrotgripp$) from the final Q-attention depth:
\begin{equation}
\label{eq:rotgrp}
\alpha, \beta, \gamma, \omega \leftarrow \argmax_{\ba^h} \qattn^{h}_{\qrotgripp}(\tilde{\vox^\qdepthN}, \ba^h) \text{ for } h \in \{0, 1, 2, 3\},
\end{equation}
where $\alpha, \beta, \gamma$ represent the individual rotation axes, $\omega$ is the gripper action, and $\tilde{\vox^\qdepthN}$ are bottleneck features from the final Q-attention depth. The final action (next-best pose) becomes $\at = (\centre^{\qdepthN}, \alpha, \beta, \gamma, \omega)$. The full system is laid out in Algorithm \ref{alg:arm}, with the main contribution highlighted in green.

\subsection{Tree Expansion}
\label{sec:tree_expansion}

The aim of the tree expansion phase is to allow for accumulation of value estimates across the top-k voxels at each Q-attention depth. We now walk through the recursive expansion, which is summarized in Algorithm \ref{alg:cte}. The $\text{QTE}$ function takes the current Q-attention depth $\qdepth$, an observation $\obs$, a voxel resolution $\res^\qdepth$, and a voxel grid center $\centre^\qdepth$. The procedure starts by first voxelizing the observations to $\vox^\qdepth$ (line 2). 
If we are not at the leaf, we run a top-k function $\topklist = \text{TOPK}_{\ba'}(\qattn_{\qattnp_{\qdepth}}(\vox^\qdepth, \ba'))$, which takes the Q-attention output at depth $\qdepth$, and returns the top-k highest Q-values (line 6).
Recall that Q-attention builds on top of Deep Q-learning~\cite{mnih2015human}, where each forward pass of the network outputs Q-values for each of the actions (voxelized locations in our case). This means that the top-k is run over the output of the network, rather than having to repeatedly run the network for each action during the top-k search.
For each of the $K$ values, we extract coordinates and set them to be the voxel grid center $\centre$ for the next depth: $\centre^{\qdepth+1} = (x_k, y_k, z_k)$ (line 9-11).
We then expand the tree at the next Q-attention depth $\qdepth+1$ by recursively calling: $\text{QTE}(\qdepth+1, \obs_t, \res^{\qdepth+1}, \centre^{\qdepth+1})$ (line 12). The output of each of these recursive calls is stored in a list, and after iterating through all $K$ branches, the max Q-value is returned (lines 13-14).

If we are at the leaf of the tree expansion (i.e., the base case of recursion), then we simply return the single highest Q-value (line 4):
\begin{equation}
\max_{\ba'} \qattn_{\qattnp_{\qdepth}}(\vox^\qdepth, \ba').
\end{equation}
To summarize, the key difference between C2F-ARM and C2F-ARM+\methAcro\ is the addition of the top-k tree expansion. Note that when $K=1$, C2F-ARM+\methAcro becomes equivalent to C2F-ARM.

Tree expansion can be integrated at two key areas within C2F-ARM: when choosing actions (for exploration) or when calculating the Q-value target (for training). Where to integrate offers different benefits: applying at action selection allows for better exploration, while using the expansion for calculating the target Q-value allows for a more accurate estimate of the value, and thus more stable gradient updates. Note that the latter is more computationally expensive than the former because of the batched operation of expanding the tree. The majority of our results applies the tree expansion for both action selection and target value calculation; we study this further in Section \ref{sec:results}.

\section{Results}
\label{sec:results}

Our results show the following 4 core takeaways:

\begin{enumerate}
    \item We are able to achieve the original performance of C2F-ARM on tasks where the tree expansion is not necessary.
    \item Our method overcomes ``coarse ambiguity'' across a set of challenging tasks involving small objects.
    \item Through a series of ablations, we study how performance varies as we adjust the tree expansion breadth $K$, and the effect of when to apply the expansion.
    \item Our real-world qualitative results also show that the advantage of the tree expansion extends to the real world. 
\end{enumerate}

\subsection{Quantitative Simulation Results}
\label{sec:simulation_results}

For our simulation experiments, we use RLBench~\cite{james2019rlbench}, which emphasizes vision-based manipulation and gives access to a wide variety of tasks with expert demonstrations. Each task has a  sparse reward of $1$, which is given only on task completion, and $0$ otherwise. We follow \textit{James et al.}~\cite{james2021coarse}, where we use a coarse-to-fine depth of $2$, each with a voxel grid size of $16^3$. Unless otherwise stated, C2F-ARM+\methAcro\ uses the tree expansion for both action selection and target Q-value selection, with $k=10$. 

\subsubsection*{\textbf{No loss in performance when using tree expansion}}
To compare C2F-ARM with and without tree expansion, we select the same 8 tasks as in previous work~\cite{james2022qattention, james2021coarse}; these are tasks that are achievable using only the front-facing camera. Additionally, \textbf{these are tasks where there is no advantage to having a tree expansion} as there is only one object in the scene, or objects can be distinguished at a coarse level due to their color. Figure \ref{fig:results_original_tasks} shows that there is no loss in performance when using \methAcro. The results illustrate that C2F-ARM+\methAcro\ achieves either similar or better performance than C2F-ARM. 

\subsubsection*{\textbf{Gain in performance on more challenging tasks}}
Unlike the previous experiment, we now purposefully choose 4 RLBench tasks where there are many objects in the scene that look similar at a coarse level. We briefly outline the tasks to emphasize why they are hard for vanilla C2F-ARM.

\begin{figure}
\centering
\includegraphics[width=1.0\linewidth]{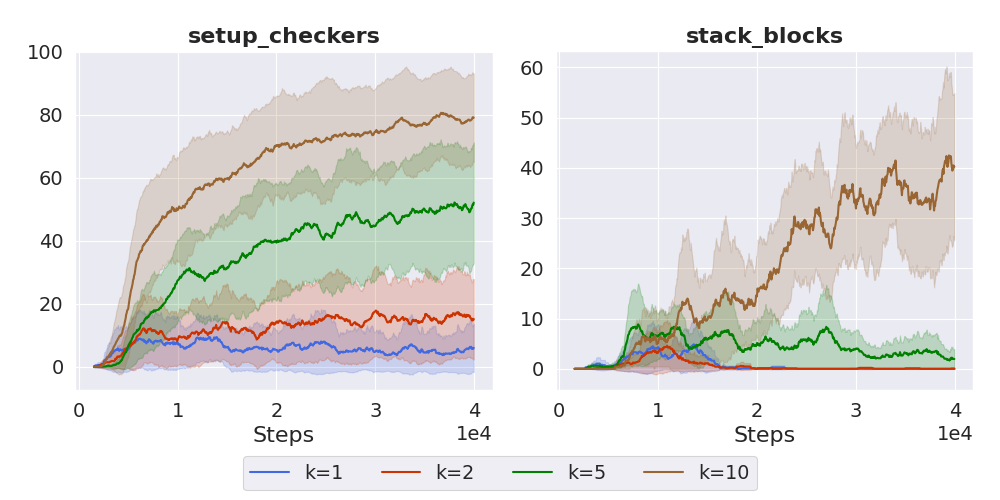}
\caption{Ablation of the tree expansion $k$ parameter on 2 RLBench tasks. Solid lines represent the average evaluation over 5 seeds, while the shaded regions represent the $std$.}
\label{fig:results_ablations_k}
\end{figure}

\begin{figure}
\centering
\includegraphics[width=1.0\linewidth]{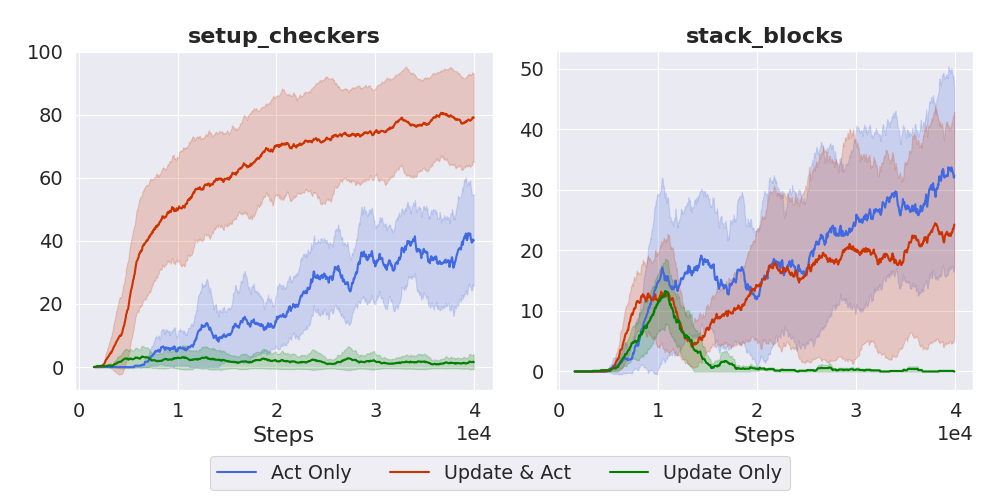}
\caption{Ablation to assess how necessary using tree expansion is for both action selection and target Q-value calculations. Solid lines represent the average evaluation over 5 seeds, while the shaded regions represent the $std$.}
\label{fig:results_ablations_update}
\end{figure}

\begin{itemize}
    \item \textit{lift\_numbered\_block}: The goal is to lift the white block with the number `1', among two distractors labeled with `2' and `3'. At a coarse level, it is not possible to read the numbers on the blocks.
    \item \textit{pick\_and\_lift\_small}: The goal is to bring the small blue cube to the target among 4 other small blue distractors with different shapes (e.g., star, moon, etc). At a coarse level, it is not possible to distinguish the subtle geometric differences between the small objects.
    \item \textit{setup\_checkers}: The goal is to complete the checkers board by placing the checker in the correct board position. At a coarse level, it is not possible to distinguish the state of the board.
    \item \textit{stack\_blocks}: The goal is to stack 2 of 4 red blocks on the target, among 4 other distractors. This is challenging because at a coarse level, each of the 4 red blocks could look equally promising to grasp, however at a finer level, it would be possible to perceive what blocks are easier to grasp first.
\end{itemize}
Note for these 4 tasks, we use a RGB-D resolution of $256 \times 256$ in order to capture all of the detail required for the task. Figure \ref{fig:results_new_tasks} presents results on these 4 tasks, and show that C2F-ARM+\methAcro\ attains better performance than C2F-ARM. There is room for improvement, especially on the \textit{lift\_numbered\_block} task; we hypothesize that this is due to the numbers still being difficult to read in the RGB image, and so further increasing the RGB-D resolution could lead to performance gains. 

\subsubsection*{\textbf{Ablations}}

\begin{figure}
\centering
\includegraphics[width=1.0\linewidth]{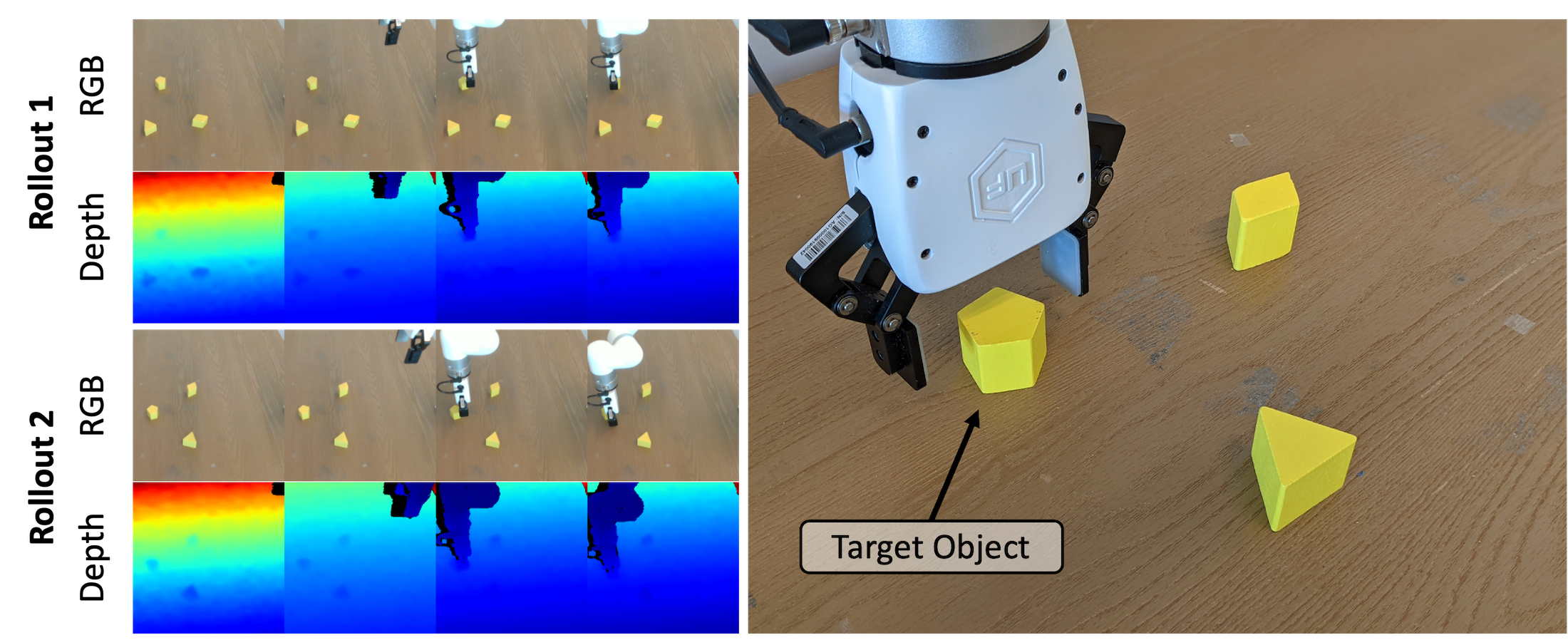}
\caption{Real world qualitative task. \textbf{Right:} The goal is to reach the Pentagonal prism, while ignoring the two distractors. Note, this is \textbf{not} the camera viewpoint. \textbf{Left:} Two rollouts from the learned agent; each column shows the RGB-D observations (from camera viewpoint) at $t=0$, $t=T/3$, $t=2T/3$ and $t=T$.}
\label{fig:real_world_reach}
\end{figure}

For our ablations, we aim to understand two key design decisions that affect performance: the effect of varying the tree expansion breadth $K$, and the effect of when to apply the expansion --- either (1) when choosing actions, (2) when calculating the Q-value target, or (3) applying both. In Figure \ref{fig:results_ablations_k}, we show the effect of varying $K$ across $1, 2, 5$ and $10$. The results show that as $K$ increases, performance improves; this makes intuitive sense, as a higher $K$ allows for more information to be brought up to the coarsest level of the Q-attention and allows for less ambiguity when choosing the highest Q-value.
In Figure \ref{fig:results_ablations_update}, we show where is best to perform the tree expansion. The results show that using tree expansion for both action selection and target Q-value selection leads to the best performance. Encouragingly, only using tree expansion for action selection also performs relatively well; this is good because the computational burden for the action selection is minimal in comparison to the target Q-value selection --- which is $\sim \times 2$ more expensive to train due to the batched tree expansion.

\subsection{Qualitative Real-world Results}
\label{sec:real_world_results}


To highlight the usefulness of the tree expansion in the real world, we design a real-world task that involves three small yellow objects with different shapes. The task is to simply reach the Pentagonal prism, while ignoring the two distractors. We give both C2F-ARM and our C2F-ARM+\methAcro\ 3 demonstrations (through tele-op via HTC Vive), and train for 10 minutes. When evaluated on 6 test episodes, C2F-ARM was only able to reach the target $2/6$ times, while C2F-ARM+\methAcro\ achieved $6/6$. Due to ``coarse ambiguity'', C2F-ARM would act as if randomly selecting one of yellow objects. For the experiments, we use the (low-cost) UFACTORY xArm 7, and a single RGB-D RealSense camera for all experiments. These qualitative results are best viewed via the videos on the project website.

\section{Conclusion} 
\label{sec:conclusion}

In this paper, we have presented our tree expansion extension to coarse-to-fine Q-attention. When the improved Q-attention is integrated into C2F-ARM --- a highly efficient manipulation algorithm --- it is able to accomplish a larger set of tasks; in particular, ones that contain small objects or objects that look similar at a coarse level. 

Below, we highlight weaknesses that would make for exciting future work. The expansion of each branch can be done independently, however our current implementation does this sequentially due to memory limits; future work could improve training efficiency by making the tree expansion parallel.
Although our method has increased the repertoire of solvable tasks, it has inadvertently introduced another hyperparameter in the from of the expansion number $K$. A large value may be unnecessary for many tasks, while for other tasks, a smaller value may cause unstable training; therefore, future work could investigate how to adapt $K$ depending on task performance. 

\section*{Acknowledgments}
This work was supported by the Hong Kong Center for Logistics Robotics.

\bibliographystyle{IEEEtran}
\bibliography{main}

\end{document}